\documentclass{ecai}

\usepackage{latexsym}
\usepackage{amssymb}
\usepackage{amsmath}
\usepackage{amsthm}
\usepackage{booktabs}
\usepackage{enumitem}
\usepackage{graphicx}
\usepackage{color}
\usepackage{subfig}
\usepackage{subfloat}
\usepackage{soul}
\usepackage[linesnumbered,ruled,vlined]{algorithm2e}
\usepackage{listings, lstautogobble}
\usepackage{rotating}
\usepackage{tabularray}
\usepackage{pifont}

\newtheorem{theorem}{Theorem}

\newtheorem{corollary}{Corollary}

\newtheorem{definition}{Definition}

\newcommand{\BibTeX}{B\kern-.05em{\sc i\kern-.025em b}\kern-.08em\TeX}

\begin{document}


\begin{frontmatter}


\paperid{5087}


\title{ToFU: Transforming How Federated Learning Systems Forget User Data}


\author[A]{\fnms{Van-Tuan}~\snm{Tran}}
\author[B]{\fnms{Hong-Hanh}~\snm{Nguyen-Le}}
\author[A]{\fnms{Quoc-Viet}~\snm{Pham}\thanks{Corresponding Author. Email: viet.pham@tcd.ie}}

\address[A]{School of Computer Science and Statistics, Trinity College Dublin, Ireland}
\address[B]{School of Computer Science, University College Dublin, Ireland}


\begin{abstract}
Neural networks unintentionally memorize training data, creating privacy risks in federated learning (FL) systems, such as inference and reconstruction attacks on sensitive data. To mitigate these risks and to comply with privacy regulations, Federated Unlearning (FU) has been introduced to enable participants in FL systems to remove their data's influence from the global model. However, current FU methods primarily act post-hoc, struggling to efficiently erase information deeply memorized by neural networks. We argue that effective unlearning necessitates a paradigm shift: designing FL systems inherently amenable to forgetting. To this end, we propose a learning-to-unlearn Transformation-guided Federated Unlearning (ToFU) framework that incorporates transformations during the learning process to reduce memorization of specific instances. Our theoretical analysis reveals how transformation composition provably bounds instance-specific information, directly simplifying subsequent unlearning. Crucially, ToFU can work as a plug-and-play framework that improves the performance of existing FU methods. Experiments on CIFAR-10, CIFAR-100, and the MUFAC benchmark show that ToFU outperforms existing FU baselines, enhances performance when integrated with current methods, and reduces unlearning time.

\end{abstract}

\end{frontmatter}


\section{Introduction}

Neural networks have been demonstrated to unintentionally memorize their training data in both centralized settings \cite{anagnostidis2022curious, bansal2022measures, carlini2022privacy} and decentralized settings \cite{thakkar2021understanding}. This memorization capability, however, poses significant privacy risks as it potentially enables the extraction of sensitive training data through various inference attacks \cite{yang2019neural, nguyen2023label} and model inversion attacks \cite{yu2023bag}. These privacy concerns are particularly pronounced in federated learning (FL) systems, where models are trained across distributed client devices containing personal data \citep{jin2021cafe, gupta2022recovering, yang2022using}. As a result, there is a growing need for mechanisms that allow the removal of private samples from trained models. This is also a requirement that is increasingly formalized in regulatory frameworks such as GDPR\footnote{https://gdpr-info.eu/art-17-gdpr/} and CCPA\footnote{https://oag.ca.gov/privacy/ccpa}.
%
In response, federated unlearning (FU) has been developed as a specialized approach, enabling participants to eradicate their data's influence from collaboratively trained models \citep{jeong2024sok, liu2023survey}. However, FU presents unique challenges compared to centralized unlearning: data remains distributed across clients, communication bandwidth is limited, the server cannot directly access client data, and client knowledge permeates throughout the global model \citep{liu2023survey}.


Despite significant progress, current FU approaches focus on modifying the model architectures or parameters to achieve unlearning objectives \citep{gu2024unlearning, halimi2022federated, liu2021federaser, wu2022federated}. These methods often overlook the fundamental relationship between learning and unlearning processes, particularly how the memorization property of neural networks impacts unlearning effectiveness. Current approaches treat the unlearning challenge as separate from the learning process, failing to address how the initial training methodology affects subsequent unlearning capabilities.

We hypothesize that improving unlearning performance requires reducing neural networks' tendency to memorize specific training instances. Instead of allowing models to memorize individual data points, we aim to guide them toward learning only task-relevant features during the training process. Note that, to enable effective unlearning in FL settings, it is crucial the reduce the permeation of knowledge during the collaborative training process \cite{liu2023survey}.
%
%
To do that, we propose incorporating a sequence of transformations applied to local data samples during training, compelling the model to learn transformation-invariant features rather than instance-specific features.
This strategy reduces explicit memorization of training data and consequently enhances unlearning performance when removal requests are processed.

The analysis on the influence of transformations on unlearning performance leads us to propose a learning-to-unlearn \textbf{T}ransf\textbf{o}rmation-guided \textbf{F}ederated \textbf{U}nlearning (ToFU) framework. ToFU integrates transformations into the federated learning process to simplify the subsequent unlearning process while preserving model utility. By establishing foundations for effective unlearning during the initial training phase, ToFU represents a paradigm shift from treating unlearning as a post-hoc \citep{halimi2022federated, liu2021federaser, wu2022federated} to designing learning systems with unlearning capabilities built in from the outset.
%
%
In particular, our framework offers key advantages: (1) \textbf{Reduced Memorization}: ToFU enforces the models to learn transformation-invariant features, substantially reducing the model's ability to memorize sample-specific information;
(2) \textbf{Seamless Versatility}: ToFU is designed to work as plug-and-play framework that improves the performance of existing FU methods;
%
(3) \textbf{Reduced Unlearning Time}: By incorporating inherent unlearning capabilities into the FL learning process, ToFU simplifies the unlearning process to a lightweight fine-tuning operation, significantly reducing computational overhead and waiting time for non-unlearning clients.

\begin{figure*}[ht]
    \centering
    \subfloat[Increasing intensity level of transformation applied to forgetting samples.]{
        \includegraphics[width=0.45\linewidth]{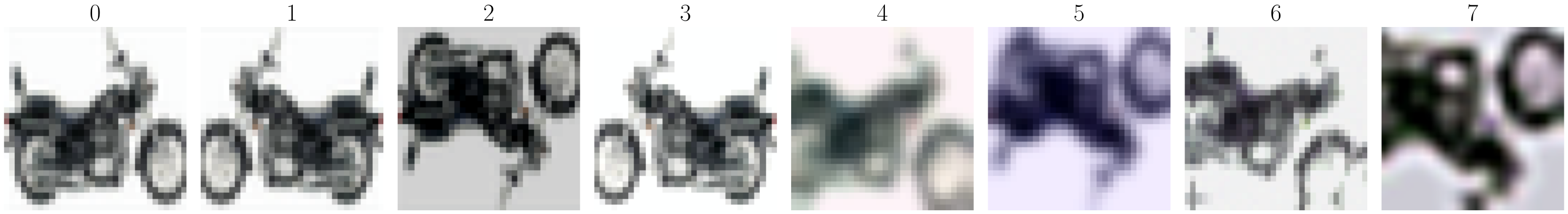}
        \label{subfig:fig:progressive_transform-examples}
    }
    \hfill
    \subfloat[Increasing masking ratio applied to forgetting samples.]{
        \includegraphics[width=0.45\linewidth]{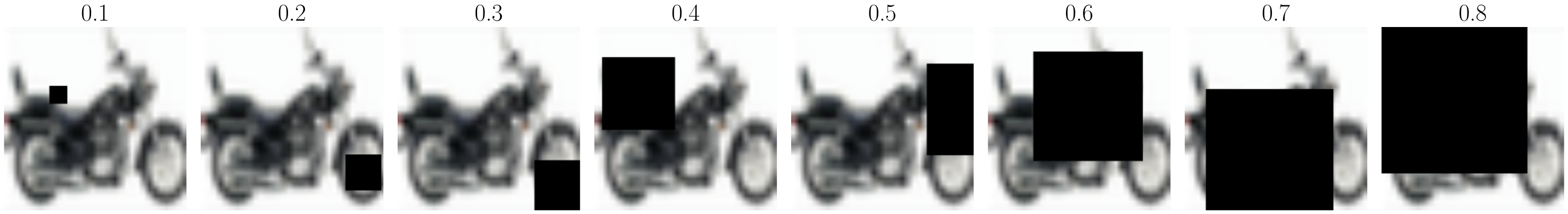}
        \label{subfig:progressive_masking-examples}
    }
    \vspace{0.05cm}
    \subfloat[Overall unlearning performance improves when increasing the intensity level of transformation on the forgetting samples.]{
        \includegraphics[width=0.45\linewidth]{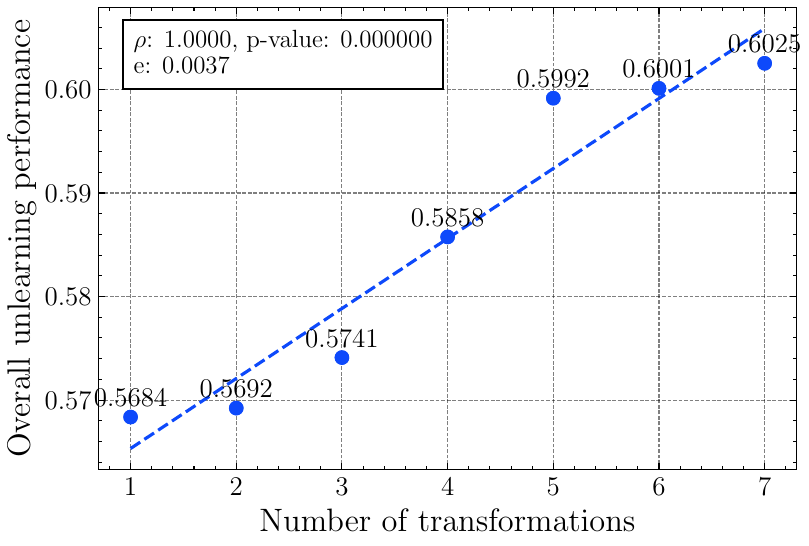}
        \label{subfig:fig:progressive_transform-correlation}
    }
    \hfill
    \subfloat[Overall unlearning performance improves when increasing the masking ratio on the forgetting samples.]{
        \includegraphics[width=0.45\linewidth]{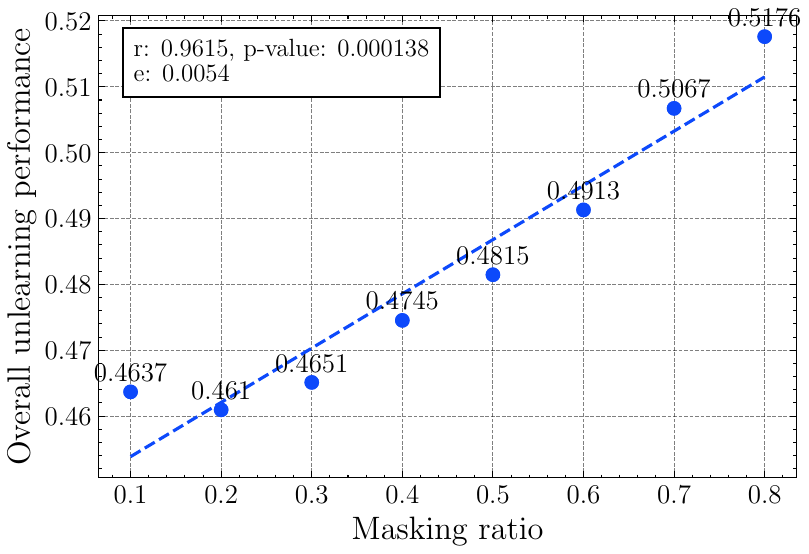}
        \label{subfig:progressive_masking-correlation}
    }
    \vspace{0.1cm}
    \caption{Experiments on CIFAR-100 to show that the unlearning process becomes more effective when models learn from transformed samples. (Left) Progressive Transformation and (Right) Progressive Masking. The top figures present forgetting samples with increasing intensity level, while the bottom figures illustrate the correlation between unlearning performance and the intensity level of transformation. $e$ is the root mean squared linear regression error, $\rho$ and $r$ are the Spearman's rank correlation and Pearson's correlation coefficients, respectively.}
    \label{fig:hypothesis-experiments}
    \vspace{0.2cm}
\end{figure*}

\textbf{Contribution.} The main contributions are summarized as
follows:
\begin{itemize}
    \item We provide theoretical and empirical results regarding the unlearning performance under a sequence of transformations of training samples. Our analysis reveals a novel insight: the unlearning process becomes more effective when models learn from transformed samples rather than memorizing raw data.
    \item We propose a learning-to-unlearn Transformation-guided Federated Unlearning (ToFU) framework that dynamically applies multiple transformations to training samples during the federated learning process. This novel approach forces the global model to learn invariant features that remain effective across all transformations of the same sample.
    \item We conduct experiments across multiple datasets, including CIFAR-10, CIFAR-100, and the MUFAC unlearning benchmark, to demonstrate the effectiveness of our approach. The results show that ToFU not only outperforms existing FU baselines but can also be integrated with current methods to enhance their performance significantly.
\end{itemize}

\section{Preliminaries}\label{sec:background}

\begin{definition}(Forget and Retain sets)
    Given a dataset $\mathcal{D} = \{\mathcal{Z}^{(i)} = (x^{(i)}, y^{(i)}) \in \mathcal{X} \times \mathcal{Y}\}_{i=1}^{N}$, $\theta^{\prime}\in \Theta$ and $f_{\theta^{\prime}}$ is the model corresponding to $\theta^{\prime}$, where $\Theta \subset \mathbb{R}^d$ be the $d$-dimensional real vector subspace. The forget, retain sets according to $\theta^{\prime}$, denoted as $\mathcal{D}_{\text{forget}}, \mathcal{D}_{\text{retain}} \subset \mathcal{D}$, is the dataset determined by $\mathcal{D}_{\text{forget}} = \{\mathcal{Z} = (x, y) \in \mathcal{D}: f_{\theta^{\prime}}(\mathcal{Z}) = False\}$ and $\mathcal{D}_{\text{retain}} = \mathcal{D} \setminus \mathcal{D}_{\text{forget}}$. Let $f_{\theta}$ denote the local model parameterized by $\theta \in \Theta$ that means $\forall \mathcal{Z}  \in \mathcal{D}_{\text{retain}}, f_{\theta}(\mathcal{Z}) = True$.
\end{definition}

\begin{definition}(Machine Unlearning)
    An unlearning process $\mathcal{U}(f_{\theta}, \mathcal{D}, \mathcal{D}_{\text{forget}})$ is defined as a transformation from a trained model $f_{\theta}$, on the dataset $\mathcal{D}$, to a unlearned model $f_{\theta^{\prime}}$ which eliminates any influence of the forget set $\mathcal{D}_{\text{forget}}$ on the $f_{\theta}$'s behavior.
\end{definition}

\begin{definition} (Federated Unlearning)
    Let $\{f_k\}_{k=1}^K$ denote a set of local models trained on local datasets $\{\mathcal{D}_k\}_{k=1}^K$ distributed among $K$ clients in a federated learning system, and let $f_{\text{global}}$ represent the global aggregate model. For any client $k \in K$ with a subset $\mathcal{D}_{k,\text{forget}} \subset \mathcal{D}_k$ designated for removal, a federated unlearning process $\mathcal{FU}(f_{\text{global}}, \{f_k\}_{k=1}^K, \{\mathcal{D}_k\}_{k=1}^K, \{\mathcal{D}_{k,\text{forget}}\}_{k=1}^K)$ is defined as a transformation from a globally trained model $f_{\text{global}}$ to a unlearned model $f_{\text{global}}^{\prime}$ which performs as if the global model had been trained with the federated learning protocol where client $k$ had only used $\mathcal{D}_{k,\text{retain}} = \mathcal{D}_k \setminus \mathcal{D}_{k,\text{forget}}$ during training, while maintaining the privacy constraints of the federated learning framework.
\end{definition}

\section{Theoretical and Empirical Analysis: Unlearning from the Lens of Transformation}\label{sec:analysis}
In this section, we establish the theoretical foundations of transformation-guided unlearning and empirically demonstrate the correlation between transformation intensity and unlearning effectiveness. We first formulate the unlearning problem through the lens of mutual information (Sec. \ref{subsec:unlearn-opt-prob}), then develop a theoretical framework showing that "\textbf{increasing the intensity level of transformation applied to forgetting samples results in better unlearning performance}" (Sec. \ref{ssec:transformation}), and finally validate our theoretical insights through experiments on benchmark datasets (Sec. \ref{ssec:empirical}).

\subsection{Unlearning Optimization Problem}\label{subsec:unlearn-opt-prob}
\begin{definition} (Mutual Information)
    For any model parameterized by $\theta$ trained on dataset $D$, we can determine the predictive distribution $p_\theta(y|x)$ for input $x$. Given two models with parameters $\theta$ and $\theta'$, the mutual information between these models with respect to a dataset $D_S$ can be defined using the Kullback–Leibler (KL) divergence:
    \begin{equation}
        I(\theta, \theta'; D_S) = D_{KL}(p_{\theta,\theta'}(x,y) \| p_\theta(x|y)p_{\theta'}(y|x)),
    \end{equation}
    where $p_{\theta,\theta'}(x,y)$ represents the joint distribution of outputs from both models, and $p_\theta(x|y)$ and $p_{\theta'}(y|x)$ represent their respective marginal distributions over dataset $D_S$.
    \label{def:mutual-info}
\end{definition}
Intuitively, mutual information measures how much information is shared between models. When $I(\theta,\theta';D_S) = 0$, the models are informationally independent with respect to $D_S$, indicating that $\theta'$ has effectively \textit{unlearned} the patterns in $D_S$ that were captured by $\theta$.
Based on this measure, we formulate the unlearning challenge as follows.

\textbf{Unlearning Optimization Problem.} Given a dataset $D$, a forget set $D_{forget}$, and $\theta$ as a trained model on $D$, find $\theta' \in \Theta$ such that:
\begin{equation}
    I(\theta, \theta'; D_{forget}) = 0 \text{ and } I(\theta, \theta'; D_{retain}) \; \text{is maximized}.
    \label{eq:unlearning}
\end{equation}
The ideal unlearning process has two goals: (1) the unlearned model should behave as if it never saw the forgotten data, and (2) it should preserve all knowledge from the retained data. Since perfectly achieving both goals simultaneously is often impossible in practice, we relax it to:
\begin{equation}
    \min I(\theta, \theta'; D_{forget}) \text{ and } \max I(\theta, \theta'; D_{retain}),
\end{equation}
which leads to the following optimization problem:
\begin{equation}
    \theta^* = \min_{\theta' \in \Theta} \frac{I(\theta, \theta'; D_{forget})}{I(\theta, \theta'; D_{retain})}.
    \label{eq:unlearning-optimization}
\end{equation}

While existing approaches \citep{gu2024unlearning, liu2021federaser, wu2022federated, halimi2022federated} assume that all forgetting samples exert equal influence on the unlearning process, we argue that applying different levels of transformations to forgetting samples can significantly improve unlearning performance.

\subsection{Transformation Composition for Improving Unlearning Performance}\label{ssec:transformation}
\begin{definition}(Transformation)
 Let $\mathcal{T}_i: \Theta \rightarrow \Theta$ be a one-shot transformation, $i = 0, 1, 2, \dots, m$, defined on parameter space $\Theta$, and $m > 0$ be a positive integer. A m-intensity transformation on $\Theta$ is a function composed of k one-shot transformations: $\mathcal{T}^{(m)} = \mathcal{T}_1 \circ \mathcal{T}_2 \circ \dots \circ \mathcal{T}_m$, where $\circ$ denotes the composition operator.
\end{definition}

We now establish how composing multiple transformations enhances unlearning effectiveness:
\begin{theorem}(Transformation Composition for Unlearning)
    Given $\mathcal{T}_0 \equiv \theta$ as a parameterized model of dataset $D$, and two simple transformations $\mathcal{T}_1 \equiv \theta'$ and $\mathcal{T}_2 \equiv \theta''$ which aim to unlearn the forget sets $D_{forget}'$ and $D_{forget}''$, respectively. Let $\mathcal{T}'^{(2)} = \mathcal{T}_1 \circ \mathcal{T}_2$ and $\mathcal{T}''^{(2)} = \mathcal{T}_2 \circ \mathcal{T}_1$. We have:
    \begin{enumerate}
        \centering
        \item $I(\theta, \mathcal{T}'^{(2)}; D_{forget}) \leq I(\theta, \mathcal{T}_1; D_{forget})$,
        \item $I(\theta, \mathcal{T}'^{(2)}; D_{forget}) \leq I(\theta, \mathcal{T}_2; D_{forget})$,
        \item  $I(\theta, \mathcal{T}''^{(2)}; D_{forget}) \leq I(\theta, \mathcal{T}_2; D_{forget})$,
        \item $I(\theta, \mathcal{T}''^{(2)}; D_{forget}) \leq I(\theta, \mathcal{T}_1; D_{forget})$,
    \end{enumerate}
    where $D_{forget} = D_{forget}' \cup D_{forget}''$.
    \label{theorem:two-transform}
\end{theorem}

Theorem \ref{theorem:two-transform} (Proof in Appendix~\ref{ssec:proof-of-theorem}) demonstrates a key insight: applying two transformations in sequence progressively on a specific sample reduces the mutual information of the model about that sample. As a result, it helps reduce the model's ability to remember specific data points, enabling the subsequent unlearning process. Each additional transformation further weakens the connection between the model and the forgotten data, regardless of the order in which transformations are applied.

\begin{corollary}(Monotonic Unlearning Performance Improvement)
    For any sequence of transformations $\mathcal{T}_1, \mathcal{T}_2, \dots, \mathcal{T}_m$ designed to unlearn different aspects of $D_{forget}$, and any composition $\mathcal{T}^{(m)} = \mathcal{T}_{1} \circ \mathcal{T}_{2} \circ \dots \circ \mathcal{T}_{m}$, we have:
    \begin{equation}
        I(\theta, \mathcal{T}^{(m)}; D_{forget}) \leq I(\theta, \mathcal{T}^{(m-1)}; D_{forget})
    \end{equation}
    \label{corollary}
\end{corollary}

This corollary (Proof in Appendix~\ref{ssec:proof-of-corollary}) follows directly from Theorem \ref{theorem:two-transform} and extends to multiple transformations. It shows that increasing the intensity level of transformation monotonically improves the unlearning performance with respect to the forget set, establishing a theoretical foundation for our transformation-guided approach.

\subsection{Empirical Validation of Transformation Effects}\label{ssec:empirical}
To validate our theoretical findings, we conduct experiments on the CIFAR-100 dataset by applying two different approaches: (1) consecutive augmentations and (2) increasing levels of masking ratio using the Cutout transformation \cite{devries2017improved}. More details about applied transformations are provided in Table \ref{tab:transformation-hyperparams} (Appendix). Figure \ref{fig:hypothesis-experiments} shows a positive correlation between the intensity level of transformation and the unlearning performance, which implies that increasing transformation intensity improves the unlearning performance. We also provide additional experiments on CIFAR-10 in the Appendix \ref{appd:CIFAR-10} that further demonstrate the role of transformation in improving unlearning performance.

\section{Transformation-guided Federated Unlearning Framework}\label{sec:FU}




In this section, we introduce the plug-and-play Transformation-guided Federated Unlearning (ToFU) framework that simplifies the unlearning process and improves the unlearning performance.


\begin{figure}[ht]
    \centering
    \includegraphics[width=0.85\linewidth]{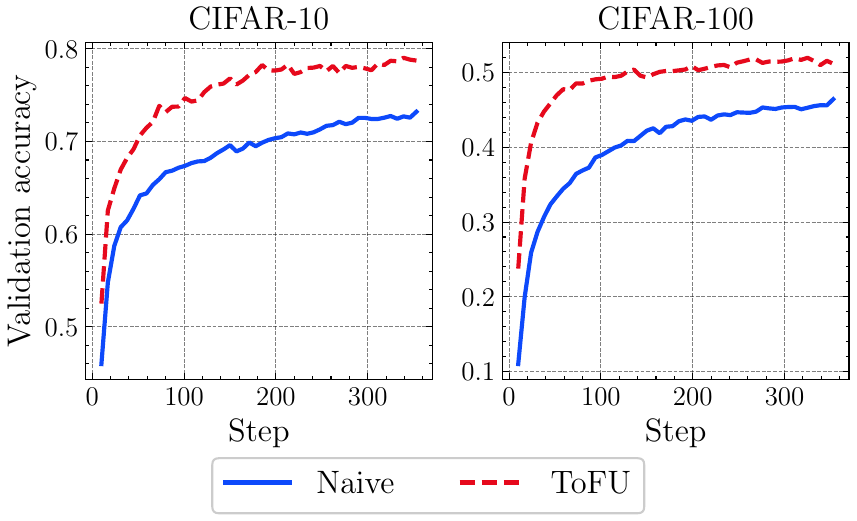}
    \caption{Learning curves of validation accuracy on CIFAR-10 and CIFAR-100 datasets. ToFU achieves better validation accuracy and faster convergence compared to the naive strategy of stacking a fixed number of transformations for every training sample.}
    \label{fig:val_learning_curves}
    \vspace{0.5cm}
\end{figure}

\textbf{Key Idea.} The fundamental principle of ToFU is that reducing memorization of specific training instances facilitates more effective unlearning. To do that, multiple transformations are dynamically applied to local training samples, inherently incorporating unlearning capabilities into the learning process itself. This means that the global model is forced to learn transformation-invariant features rather than memorizing sample-specific characteristics. However, training with extensively transformed data presents significant challenges, including potential distortion of task-relevant information \cite{tian2020makes, wang2022contrastive} that can lead to underfitting and convergence issues, as demonstrated in Figure \ref{fig:val_learning_curves}. To mitigate this, we introduce a \textit{sample-dependent transformation strategy} to dynamically determine the optimal intensity level of transformation for each sample. Furthermore, we also propose a \textit{transformation-invariant regularization technique} to encourage the model to exploit invariant features across all transformations of the same sample. The complete algorithm is outlined in Algorithm \ref{alg:ToFU}, while types of transformations used in our experiments are provided in Algorithm \ref{alg:tofu_transformations} (Appendix).

\textbf{Sample-dependent Transformation Strategy.} While applying multiple transformations helps reduce pattern memorization, excessive transformations can eliminate task-relevant information and degrade performance \cite{tian2020makes}. It is crucial to determine how much information is shared about the input, including both task-relevant and irrelevant information. In other words, we need to find the optimal intensity level of transformation that can be applied to per sample. Particularly, we introduce an inverse quantile value function (Definition \ref{def:inverse-quantile-value}) that dynamically determines the appropriate intensity level of transformation for each sample. This function leverages per-sample loss values calculated using the latest global model $f_\theta^{t-1}$ during each round $t$. It establishes an inverse relationship between sample loss and transformation intensity: samples with lower loss values (i.e., less irrelevant information) receive more transformations.

Furthermore, we implement a progressive training approach by linearly increasing the maximum intensity level of transformation $M$ throughout the training process $M := \lceil \frac{t}{T} M \rfloor$, where $T$ denotes the total number of communication rounds and $t$ represents the current round. For each training mini-batch, ToFU creates a transformed mini-batch with sample-specific transformation counts determined by relative loss values within the batch.



\begin{definition}
\label{def:inverse-quantile-value}
Let $\mathbf{x} = (x_1, x_2, \ldots, x_n)$ be a vector of batch loss values.
The inverse quantile value function $\digamma: \mathbb{R}^n \rightarrow [0, 1]^n$ is defined as:
$$
\digamma(\mathbf{x}) = \left(\frac{\sum_{j=1}^n \mathbf{1}_{x_j > x_i}}{n}\right)_{i=1}^n,
$$
where $\mathbf{1}$ denotes the indicator function.
The number of applied transformations $A_i$ for sample $i$ is then determined by:
$$
A_i = \left\lceil M \cdot \digamma(\mathbf{x})_i \right\rceil,
$$
where $M$ is the maximum number of allowed transformations.
\end{definition}

\textbf{Transformation-invariant Regularizer.} The transformation-invariant regularizer serves a critical function in our framework by compelling the model to identify and exploit correlated structures across multiple transformations rather than memorizing individual data points.
%
%
Particularly, beyond the standard supervised learning signal, we incorporate the KL divergence between the latent representations of transformed data $f_{\theta}(\mathcal{T}(\mathcal{D}))$ and that of original data $f_{\theta}(\mathcal{D})$, where $\mathcal{T}$ denotes the transformation function. This regularization term maximizes mutual information between representations of transformed and original data, encouraging consistent feature extraction across different versions of the same sample. Formally, for each sample-label pair $(x, y)$ in the client's dataset, the loss function for progressive training is defined as follows:
\begin{equation}
\ell(\theta; x, y) = \underbrace{-y\log(p_\theta(x^*))}_{\text{Task loss}} + \gamma \underbrace{D_{KL}(f_\theta(x^*) \parallel f_\theta(x))}_{\text{Transformation-invariant reg.}},
\label{eq:loss_function}
\end{equation}
In our experiments, we use $\gamma=0.01$ unless stated otherwise.

\begin{algorithm}[ht]
\small
\DontPrintSemicolon
\SetKwInOut{Input}{Input}
\SetKwInOut{Output}{Output}
\SetKwFunction{LocalTraining}{LocalTraining}
\SetKwFunction{SampleTransform}{SampleTransform}
\SetKwComment{Comment}{$\triangleright$ }{}
\SetKwProg{Fn}{Function}{:}{}

\Input{Number of clients $K$, batch size $B$, learning rate $\eta$, epochs $E$, max transformation level $M$}
\Output{Trained global model $\theta_T$ with improved unlearning capabilities}
\textbf{\underline{Server-side Coordination}} \;
Initialize global model $\theta_0$ \;
\For{communication round $t = 1, 2, \ldots, T$}{
    \For{each client $k \in S_t$ in parallel}{
        $\theta_{t+1}^k \gets$ \LocalTraining{$k, \theta_t$} \;
    }
    $\theta_{t+1} \gets \sum_{k=1}^K \frac{|D_k|}{\sum_{j=1}|D_j|} \theta_{t+1}^k$ \Comment{Federated aggregation}
}
\textbf{\underline{Client-side Training}} \;
\Fn{\LocalTraining{$k, \theta$}}{
    \For{each epoch from $1$ to $E$}{
        \For{each batch $\mathcal{B}$ sampled from client dataset $D_k$}{
            $\mathcal{B} \gets \{(x_1, y_1), \ldots, (x_B, y_B)\}$ \;
            $\mathcal{L}(\theta) \gets \ell(\theta; \mathcal{B})$ \Comment{Calculate batch loss, stop-gradient}
            $\mathcal{B}^* \gets$ \SampleTransform{$\mathcal{B}, \mathcal{L}, M$} \Comment{Apply transformations}
            \Comment{Transformation-invariant regularization}
            $\mathcal{L}^*(\theta) \gets \ell(\theta; \mathcal{B^{*}}) + \gamma \mathrm{KL}(f_\theta(\mathcal{B}^*), f_\theta(\mathcal{B}))$ \;
            $\theta \gets \theta - \eta \nabla_\theta \mathcal{L}^*(\theta)$ \Comment{Gradient descent}
        }
    }
    \KwRet{$\theta$} to server \;
}
\textbf{\underline{Sample-dependent Transformation}} \;
\Fn{\SampleTransform{$\mathcal{B}, \mathcal{L}, M$}}{
    \For{each sample $(x_i, y_i)$ in $\mathcal{B}$}{
        \Comment{Determine transformation intensity based on batch loss}
        $m \gets \left\lceil\frac{|\{j : \mathcal{L}_j > \mathcal{L}_i\}|}{|\mathcal{B}|} \cdot M\right\rceil$ \;
        $x_i^* \gets \mathcal{T}^{(m)}(x_i)$ \Comment{Apply $m$-intensity transformation}
    }
    \KwRet{$\mathcal{B}^* \gets \{(x_1^*, y_1), \ldots, (x_B^*, y_B)\}$} \;
}
\textbf{\underline{Unlearning Procedure}} \;
\For{each client $k$ with forget request}{
    \For{each epoch from $1$ to $E_u$}{
        \For{each batch $\mathcal{B}$ sampled from $D_{k_{retain}}$}{
          $\mathcal{L}(\theta) = \ell(\theta; \mathcal{B})$ \;
          $\theta = \theta - \eta_u \nabla_{\theta}\mathcal{L}(\theta)$
        }
    }
}

\caption{ToFU: Transformation-guided Federated Unlearning}
\label{alg:ToFU}
\end{algorithm}

\textbf{Unlearning Procedure.} From Alg. \ref{alg:ToFU}, when a removal request is issued, the unlearning phase involves a simple fine-tuning stage on the retain set. This shows that our method helps simplify the unlearning process.

\section{Experiments}
\subsection{Experimental Setup}
\textbf{Datasets.} We use two canonical datasets, CIFAR-10 and CIFAR-100 \cite{krizhevsky2009learning}, and the unlearning-specific dataset, MUFAC \cite{choi2023towards}. The client's data distribution is sampled from a $\text{Dirichlet}$ distribution, with $p_{k,c} \sim \mathfrak{D}(\kappa)$, using a concentration parameter $\kappa = 1.0$. Here, $p_{k, c}$ represents the proportion of samples from class $c$ assigned to client $k$. For the forgetting samples, we randomly select $10\%$, $30\%$, $60\%$, and $90\%$ of data from clients $1$, $3$, $6$, $9$, respectively. For MUFAC, we use the pre-defined forget set as described in the original paper \cite{choi2023towards}.

\textbf{Metrics.} We use three metrics, including Retain Accuracy, Test Accuracy, and Membership Inference Attack Efficacy (MIA Efficacy). For MIA Efficacy, we employ the state-of-the-art (SoTA) method - LiRA \cite{carlini2022membership}. Details about these metrics are provided in Section \ref{sec:implementation-details} (Appendix).

\textbf{Baselines.} Regarding the FU baselines, we consider three methods: (1) \textbf{FedEraser} \cite{liu2021federaser}; (2) The FU method using Projected Gradient Descent (PGD) (\textbf{FedPGD}) by \cite{halimi2022federated}; (3) The FU method using the AdaHessian optimizer (\textbf{FedAda}) \cite{liu2022right}. To provide a comprehensive evaluation, we also include two additional baselines: an exact unlearning baseline (retrain-from-scratch), which is the gold standard of unlearning, and a decentralized adaptation of an MU method that employs $\ell_1$-sparsification followed by fine-tuning \cite{liu2024model}.

\textbf{Implementation.} In the collaborative learning stage, we conduct the FL simulations over $T=50$ communication rounds for $K = 10$ clients’ models.
%
%
The unlearning stage applies FU methods for $20$ rounds and the number of local epochs is set to $5$. More implementation details are provided in Section \ref{sec:implementation-details} (Appendix).

\vspace{0.5cm}
\begin{table}[ht]
\centering
\fontsize{7pt}{7pt}\selectfont
\caption{Comparison with FU methods, including FedEraser, FedPGD, and FedAda on CIFAR-10, CIFAR-100, and MUFAC datasets. $\uparrow$ means higher values are better for all metrics.}
\label{tab:tofu-outperform}
\begin{tblr}{
  width = \linewidth,
  colspec = {Q[75]Q[160]Q[110]Q[110]Q[110]Q[80]},
  cells = {c},
  cell{1}{1} = {r=2}{},
  cell{1}{2} = {r=2}{},
  cell{1}{3} = {c=4}{0.1\linewidth},
  cell{3}{1} = {r=4}{},
  cell{7}{1} = {r=4}{},
  cell{11}{1} = {r=4}{},
  vline{2-3} = {1-2}{},
  vline{2-3} = {3-14}{},
  vline{6} = {2-14}{},
  hline{1,15} = {-}{0.08em},
  hline{2} = {3-6}{},
  hline{3,7,11} = {-}{},
  hline{3} = {2}{-}{},
}
\textbf{Dataset}                        & \textbf{Method} & \textbf{Metrics}$\uparrow$ &                   &                       &                  \\
                                        &                 & \textbf{Test Accuracy}         & \textbf{Retain Accuracy} & \textbf{MIA Efficacy} & \textbf{Overall} \\
\begin{sideways}CIFAR-10\end{sideways}  & FedEraser \cite{liu2021federaser}      & 0.7685                     & 0.8739            & 0.2926                & 0.6450           \\
                                        & FedPGD \cite{halimi2022federated}         & 0.7826                     & 0.8959            & 0.2910                & 0.6565           \\
                                        & FedAda \cite{liu2022right}         & 0.7755                     & 0.8754            & 0.2891                & 0.6466           \\
                                        & \textbf{ToFU}   & \textbf{0.7943}            & \textbf{0.8955}   & \textbf{0.3239}       & \textbf{0.6712}  \\
\begin{sideways}CIFAR-100\end{sideways} & FedEraser \cite{liu2021federaser}      & 0.4764                     & 0.7425            & 0.2949                & 0.5046           \\
                                        & FedPGD \cite{halimi2022federated}         & 0.4682                     & 0.6659            & 0.2221                & 0.4520           \\
                                        & FedAda \cite{liu2022right}         & 0.4803                     & 0.6920            & 0.3629                & 0.5117           \\
                                        & \textbf{ToFU}   & \textbf{0.5032}            & \textbf{0.7802}   & \textbf{0.4560}       & \textbf{0.5798}  \\
\begin{sideways}MUFAC\end{sideways}     & FedEraser \cite{liu2021federaser}      & 0.8597                     & 0.8900            & 0.3586                & 0.7027           \\
                                        & FedPGD \cite{halimi2022federated}         & 0.8431                     & 0.8776            & 0.3757                & 0.6988           \\
                                        & FedAda \cite{liu2022right}         & 0.7600                     & 0.8158            & 0.3934                & 0.6564           \\
                                        & \textbf{ToFU}   & \textbf{0.8943}            & \textbf{0.9379}   & \textbf{0.4651}       & \textbf{0.7657}
\end{tblr}
\end{table}

\begin{table}[ht]
\centering
\fontsize{7pt}{7pt}\selectfont
\caption{Comparison with exact unlearning and MU methods. $\uparrow$ means higher values are better for all metrics.}
\label{tab:comparison_mu_exact_unlearning}
\begin{tblr}{
  width = \linewidth,
  colspec = {Q[70]Q[180]Q[110]Q[110]Q[100]Q[70]},
  cells = {c},
  cell{1}{1} = {r=2}{},
  cell{1}{2} = {r=2}{},
  cell{1}{3} = {c=4}{0.1\linewidth},
  cell{3}{1} = {r=3}{},
  cell{6}{1} = {r=3}{},
  cell{9}{1} = {r=3}{},
  vline{2-3} = {1-2}{},
  vline{2-3} = {3-11}{},
  vline{6} = {2-11}{},
  hline{1,12} = {-}{0.08em},
  hline{2} = {3-6}{},
  hline{3,6,9} = {-}{},
  hline{3} = {2}{-}{},
}
\textbf{Dataset}                        & \textbf{Method}                       & \textbf{Metrics$\uparrow$} &                   &                       &                  \\
                                        &                                       & \textbf{Test Accuracy}         & \textbf{Retain Accuracy} & \textbf{MIA Efficacy} & \textbf{Overall} \\
\begin{sideways}CIFAR-10\end{sideways}  & Exact unlearning                  & 0.7515                     & 0.8464            & 0.4645                & \textbf{0.6874}           \\
                                        & $\ell_1$-sparsity~\cite{liu2024model} & 0.6082                     & 0.6045            & \textbf{0.4653}       & 0.5593           \\
                                        & \textbf{ToFU}                         & \textbf{0.7943}            & \textbf{0.8955}   & 0.3239                & 0.6712  \\
\begin{sideways}CIFAR-100\end{sideways} & Exact unlearning                  & 0.4584                     & 0.5729            & 0.5007                & 0.5106           \\
                                        & $\ell_1$-sparsity~\cite{liu2024model} & 0.4084                     & 0.4630             & \textbf{0.6124}       & 0.4946           \\
                                        & \textbf{ToFU}                         & \textbf{0.5032}            & \textbf{0.7802}   & 0.4560                & \textbf{0.5798}  \\
\begin{sideways}MUFAC\end{sideways}     & Exact unlearning                  & 0.5146                     & 0.8900              & 0.5375                & 0.6473           \\
                                        & $\ell_1$-sparsity \cite{liu2024model} & 0.1809                     & 0.2242            & \textbf{0.8603}       & 0.4218           \\
                                        & \textbf{ToFU}                         & \textbf{0.8943}            & \textbf{0.9379}   & 0.4651                & \textbf{0.7657}
\end{tblr}
\end{table}

\subsection{Experimental Results}
In this section, we evaluate ToFU across four experimental settings: comparison with existing FU methods, integration with current approaches, benchmarking against exact unlearning and machine unlearning (MU) baselines, and performance verification in non-unlearning scenarios.

\subsubsection{Comparison with FU methods}
To show the performance of ToFU, we compare it with existing FU works: FedEraser, FedPGD, and FedAda. Table \ref{tab:tofu-outperform} shows that our ToFU framework outperforms existing FU methods across all datasets.
Notably, while not only being superior in model utility metrics (Test and Retain accuracies), ToFU also shows better privacy preservation via the efficacy against MIA attacks. We provide a detailed analysis in Section \ref{sssec:analysis-MIA} to highlight this effectiveness.

\subsubsection{Performance improvement over existing FU methods} {Beyond outperforming existing FU methods, ToFU also demonstrates its versatility as it can be effectively integrated with existing FU frameworks to further enhance their performance. As illustrated in Figure \ref{fig:tofu-improvements}, when incorporated with FedEraser, FedPGD, and FedAda, ToFU consistently improves its performance across all datasets. Specifically, ToFU improves their overall FU performance by significant margins of $6.030 \pm{1.359}\%$, $11.007 \pm{1.902}\%$, $4.733 \pm{1.346}\%$, respectively. These results highlight that ToFU can significantly boost the performance of other FU methods through its sample-dependent transformation strategy.

\begin{figure*}[ht]
    \centering
    \includegraphics[width=0.8\textwidth]{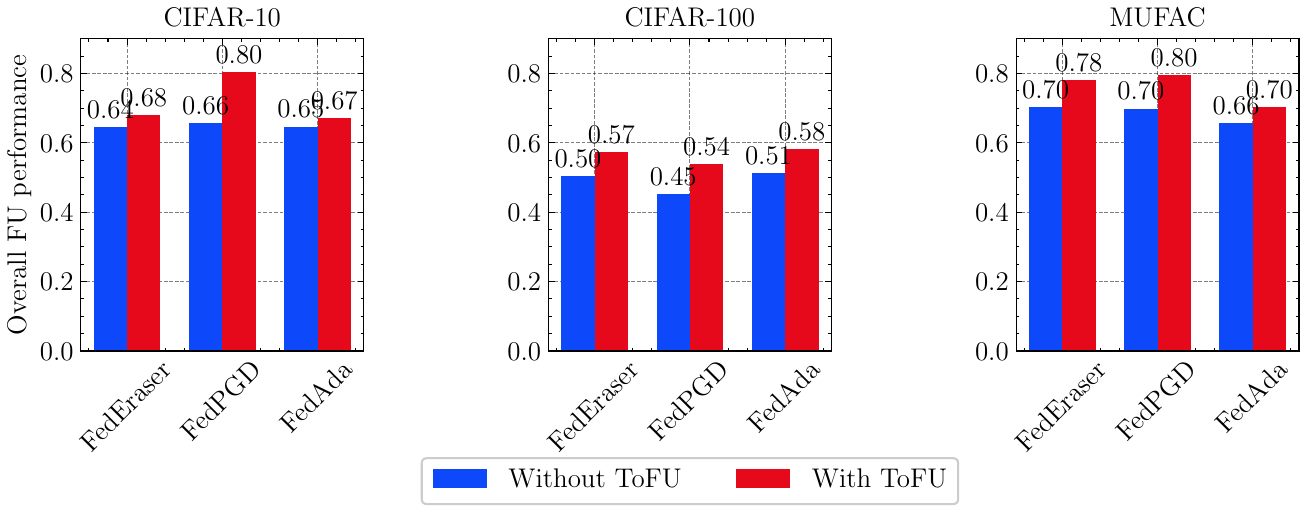}
    \caption{Illustration of the performance improvement of existing FU methods when being incorporated with ToFU across three datasets: CIFAR-10, CIFAR-100, and MUFAC.}
    \label{fig:tofu-improvements}
    \vspace{0.25cm}
\end{figure*}

\begin{figure*}[ht]
\centering
\includegraphics[width=0.9\textwidth]{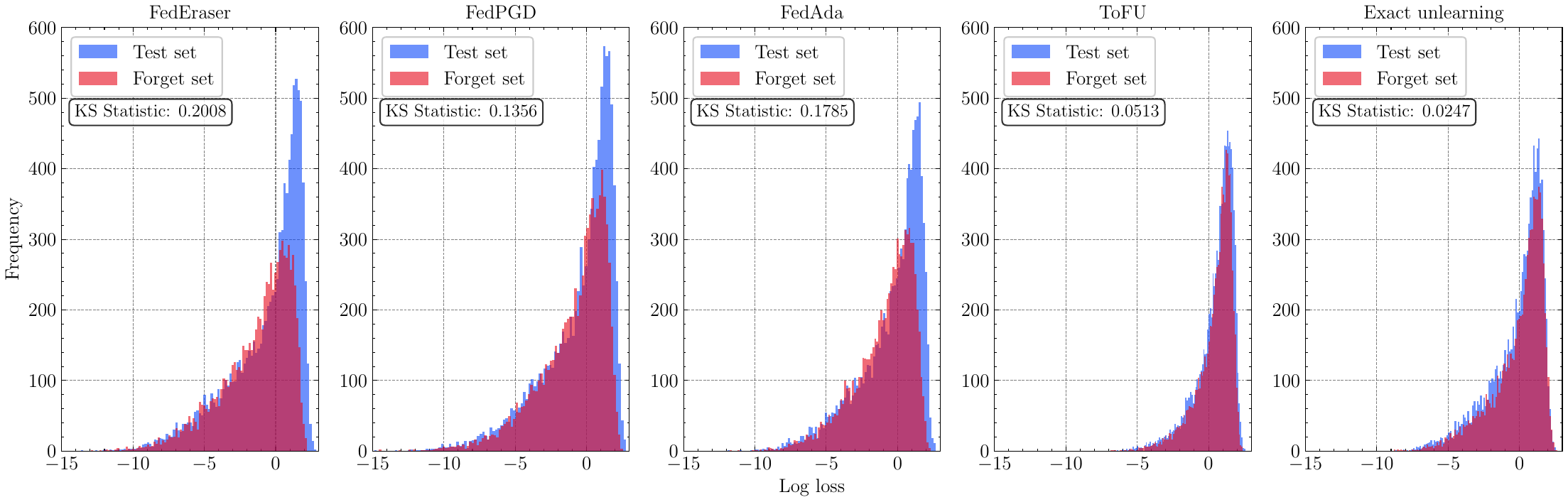}
\caption{The histogram of loss value of unlearned models on the test set (unseen data) and the forget set of CIFAR-100, comparing FU approaches: FedEraser, FedPGD, FedAda, ToFU, and the gold standard exact unlearning. We also provide the Kolmogorov-Smirnov (KS) statistics to quantitatively measure the divergence between distributions of the outputs of the unlearned model on the test set and the forget set.}
\label{fig:tofu-improvements-MIA-loss}
\vspace{0.1cm}
\end{figure*}

\subsubsection{Comparison with exact unlearning and decentralized MU} In this experiment, we further compare our ToFU framework with the exact unlearning baseline and a decentralized adaptation of the MU method \cite{liu2024model}.
From Table \ref{tab:comparison_mu_exact_unlearning}, we observe that:
\begin{itemize}
    \item ToFU achieves overall performance scores of $0.6712$, $0.5798$, and $0.7657$ on CIFAR-10, CIFAR-100, and MUFAC, respectively. ToFU's performance is comparable to that of the exact unlearning baseline, which is widely considered the gold standard for unlearning \cite{liu2023survey}.
    \item Although $\ell_1$-sparsity~\cite{liu2024model} effectively mitigates the vulnerability to MIAs and obtains SOTA performance in MU, its sparsification strategy is not fully compatible with FU settings, thus compromising model utility, as reflected in lower test and retain accuracies. In contrast, ToFU successfully strikes a balance of FU performance from both utility and privacy perspectives.
\end{itemize}

\begin{table}[t]
\centering
\fontsize{7pt}{7pt}\selectfont
\caption{FL performance of FedAvg when training with and without ToFU. $\uparrow$ means higher values are better.}
\label{tab:analysis_extreme_case}
\begin{tblr}{
  cells = {c},
  cell{1}{1} = {r=2}{},
  cell{1}{2} = {c=2}{},
  vline{2} = {1-5}{},
  hline{1,6} = {-}{0.08em},
  hline{2} = {2-3}{},
  hline{3} = {-}{},
  hline{3} = {2}{-}{},
}
\textbf{Dataset} & \textbf{Test Accuracy}$\uparrow$ &                      \\
                 & \textbf{FedAvg}          & \textbf{FedAvg + ToFU} \\
CIFAR-10         & 0.7803                           & \textbf{0.7843}      \\
CIFAR-100        & 0.5013                           & \textbf{0.5189}      \\
MUFAC            & \textbf{0.8451}                  & 0.8165
\end{tblr}
\end{table}

\subsubsection{Performance of ToFU in non-unlearning scenario}
We examine ToFU in an extreme FU scenario where no clients request data removal. Table \ref{tab:analysis_extreme_case} demonstrates that ToFU maintains performance integrity even in this edge case, showing no detrimental effects on the overall performance.
For canonical, natural image datasets like CIFAR-10 and CIFAR-100, ToFU outperforms the vanilla FedAvg baseline, due to the implicit regularization effects of transformations \cite{lejeune2019implicit}.
Conversely, on the MUFAC unlearning benchmark dataset, which contains real-world facial images, the transformations used by ToFU marginally decrease the test accuracy from $0.8451$ to $0.8165$.
This slight performance degradation occurs because MUFAC contains real facial images with varying lighting conditions, making it inherently less amenable to transformations compared to natural image datasets like CIFAR-10/100.
This outcome is consistent with existing research highlighting the effects of transformation on different datasets and the necessity for tailored transformation strategies in specialized tasks \cite{balestriero2022effects, tian2020makes}.

\subsection{Analysis and Ablation Study}
Having established ToFU's performance advantages, we now examine ToFU's improved robustness against membership inference attacks and evaluate its computational efficiency compared to existing FU methods. Moreover, we also provide an ablation study to evaluate individual contributions of ToFU's components. Other ablation studies on transformation-invariant regularize strengths and wall-clock time for training are provided in Sec. \ref{sec:additional-ablation} (Appendix).

\subsubsection{Improved robustness against membership inference attacks}\label{sssec:analysis-MIA}
We conduct experiments with the LiRA attack \cite{carlini2022membership} to show the robustness of ToFU framework.
Given that most MIAs rely on loss-based inference \cite{shokri2017membership, carlini2022membership}, we argue that minimizing the discrepancy between the loss value distribution of forgetting samples and unseen samples (i.e., test set) is crucial for effective unlearning. Through data transformations and mutual information maximization, ToFU achieves this by not directly revealing the actual forgetting samples during training, thereby reducing the likelihood of memorization and susceptibility to membership inference. Our analysis, as illustrated in Figure \ref{fig:tofu-improvements-MIA-loss}, demonstrates that ToFU effectively narrows the divergence of loss values on the forget and the test sets.

We further conduct the Kolmogorov–Smirnov (KS) statistical test to quantify the difference between the output distributions of the unlearned models on the test set and the forget set. Figure \ref{fig:tofu-improvements-MIA-loss} shows that ToFU not only more effectively reduces the statistical divergence between these two distributions compared to other FU methods but also achieves a level of distributional similarity on par with the gold standard exact unlearning.

\begin{figure}[ht]
    \centering
    \includegraphics[width=0.7\linewidth]{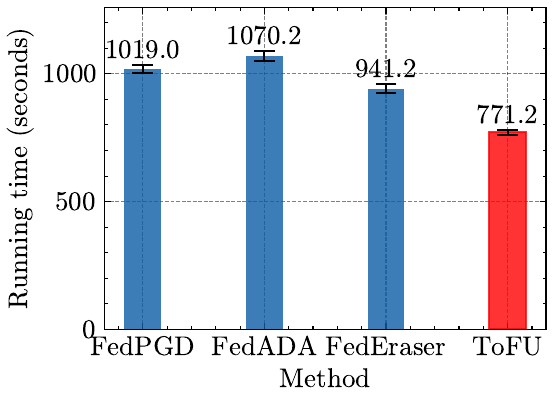}
    \caption{Running time of ToFU compared to other FU methods.}
    \label{fig:running}
    \vspace{0.4cm}
\end{figure}

\subsubsection{Running time of ToFU}
This section examines the running-time efficiency of ToFU in the unlearning process. Figure \ref{fig:running} illustrates the significant computational advantage offered by our TFU framework compared to existing FU approaches. With a mean running time of $771.2$ seconds, TFU outperforms FedPGD, FedADA, and FedEraser, representing a $24-28\%$ reduction in running time. This efficiency stems directly from TFU's core design: by incorporating transformations during initial training, the framework inherently simplifies subsequent unlearning processes.

\begin{table}[ht]
\centering
\small 
\caption{Ablation study on ToFU components.}
\label{tab:components}
\begin{tblr}{
  width = \linewidth,
  colspec = {X[c] X[c] X[c] X[c] X[c] X[c] X[c]},
  cells = {c},
  vline{2-7} = {-}{},
  hline{1,8} = {-}{0.1em},
  hline{2} = {-}{},
  hline{2} = {2}{-}{},
  rowsep=2pt
}
\textbf{SDT}\TblrNote{a} & \textbf{TIR}\TblrNote{b} & \textbf{PT}\TblrNote{c} & \textbf{Test Acc.} & \textbf{Retain Acc.} & \textbf{MIA Eff.} & \textbf{Overall} \\
\ding{51}                &                          &                         & 0.4790            & 0.6414              & 0.5699           & 0.5634           \\
\ding{51}                & \ding{51}                &                         & 0.4880            & 0.6430              & 0.5851           & 0.5720           \\
\ding{51}                &                          & \ding{51}               & 0.5124            & 0.8124              & 0.3394           & 0.5547           \\
                         & \ding{51}                &                         & 0.4995            & 0.9231              & 0.1848           & 0.5358           \\
                         & \ding{51}                & \ding{51}               & 0.4963            & 0.9223              & 0.1903           & 0.5363           \\
\ding{51}                & \ding{51}                & \ding{51}               & {0.5032}          & {0.7802}            & {0.4560}         & \textbf{0.5798}
\end{tblr}
\noindent\hspace*{\hangindent}\textit{Note:} \TblrNote{a} SDT: Sample-dependent Transformation.
\TblrNote{b} TIR: Transformation-invariant Regularizer.
\TblrNote{c} PT: Progressive Training.
\end{table}

\subsubsection{Ablation study on ToFU's components}
This ablation study evaluates the individual and combined contributions of ToFU's three core components: Sample-dependent Transformation, Transformation-invariant Regularizer, and Progressive Training. As shown in Table \ref{tab:components}, integrating all three components achieves the highest average performance ($0.5798$) and the most balanced performance across metrics: Test Accuracy ($0.5032$), Retain Accuracy ($0.7802$), and MIA Efficacy ($0.4560$).

\section{Related Works}
\label{sec:related-work}
\subsection{Machine Unlearning}
Research in the field of MU has focused on developing efficient methods to eliminate the influence of specific data points or subsets from trained models without requiring complete retraining. Common approaches include model scrubbing \citep{golatkar2020eternal}, and leveraging model sparsity \citep{liu2024model}.

Another line of work has aimed to understand the fundamental limits, trade-offs, and properties of unlearning algorithms through theoretical analysis. These studies provide a probabilistic definition of MU \citep{ginart2019making}, certified data removal for unlearning algorithms \citep{izzo2021approximate}, and a theoretical upper bound for MU scalability \citep{sekhari2021remember}.

Furthermore, some studies have applied MU techniques to specific domains, such as large language models \citep{eldan2023s} and image generation models \citep{fan2023salun}, demonstrating the utility of these methods in various real-world scenarios.

\subsection{Federated Unlearning}
Federated Unlearning (FU) has emerged as a critical mechanism to enable clients' right to be forgotten in FL systems. Existing FU methodologies can be categorized into two approaches: \textit{server-side} and \textit{client-side} unlearning. Server-side unlearning techniques focus on modifying the central server's global model to eliminate specific clients' influence. \citet{liu2021federaser} first proposed storing historical client updates on the server for post-hoc recalibration. However, raises significant privacy concerns due to the retention of client updates, incurs substantial storage overhead, and suffers from calibration errors during the unlearning process. Subsequent works attempted to mitigate these issues through knowledge distillation \citep{wu2022federated} and auxiliary datasets \citep{fu2024client}.

On the other hand, client-side unlearning leverages local clients' direct access to their forget and retain sets for local model scrubbing, which enables unlearning at the instance level.
\citet{wang2022federated} \citep{wang2022federated} specifically designed a pruning-based approach for Convolutional Neural Networks (CNNs) which uses channel pruning guided by Term Frequency - Inverse Document Frequency.
Other works have introduced more general FU methods, such as those utilizing gradient ascent for model scrubbing \citep{halimi2022federated} and variational Bayesian inference for forgetting and memorizing balance \citep{wang2023bfu}. Recent developments in FU research have expanded to explore additional critical aspects, such as the verification mechanism of unlearning \citep{gao2024verifi}.
Particularly, a framework named VERIFI \citep{gao2024verifi} was introduced to provide the unlearning clients the right to verify the FU process.

Compared to previous works, our ToFU framework represents a paradigm shift from treating unlearning as post-hoc to designing learning systems with built-in unlearning capabilities. Particularly, ToFU incorporates transformation-guided learning during the federated training process itself, which helps simplify the subsequent unlearning processes and improve the unlearning performance.

\section{Conclusions}
In this work, we have theoretically and empirically demonstrated that the unlearning performance can be improved by incorporating a sequence of transformations into training samples in the federated learning process. We have presented a Transformation-guided Federated Unlearning (ToFU) framework that establishes a new paradigm that makes neural networks inherently more amenable to unlearning. Our experiments across CIFAR-10, CIFAR-100, and MUFAC benchmark datasets have demonstrated that ToFU not only outperforms existing FU methods in terms of privacy protection against membership inference attacks, but also maintains high model utility on retained and test data. Furthermore, ToFU's compatibility with existing unlearning approaches and its significant improvement in unlearning time reduction ($24-28\%$) make it a practical solution for real-world privacy-preserving FL systems. As regulations increasingly mandate the right to be forgotten, our transformation-guided approach provides a promising direction for building FL systems that respect privacy by design rather than as an afterthought.

In practice, to enhance the generalizability of ToFU to different transformation families, the automatic augmentation policy, RandAugment \cite{cubuk2020randaugment}, can be used. This technique has been proven to generalize well across diverse architectures and datasets without extensive hyperparameter tuning.


\break \newpage
\begin{ack}
This publication has emanated from research conducted with the financial support of Taighde Éireann – Research Ireland under Grant number 18/CRT/6222. For the purpose of Open Access, the author has applied a CC BY public copyright licence to any Author Accepted Manuscript version arising from this submission.
The work of Quoc-Viet Pham is supported in part by European Union under the ENSURE-6G project (Grant Agreement No. 101182933).
\end{ack}




\bibliography{main}

\clearpage
\appendix
\section*{Appendix for ToFU: Transforming How Federated Learning Systems
Forget User Data}\label{sec:appendix}

This appendix provides supplementary materials to support the main paper. It includes detailed proofs for Theorem \ref{theorem:two-transform} and Corollary \ref{corollary} (Section \ref{sec:proofs}), additional empirical results on the CIFAR-10 dataset demonstrating the effectiveness of transformation-guided unlearning (Section \ref{appd:CIFAR-10}), implementation details (Section \ref{sec:implementation-details}), transformations details (Section \ref{sec:transform-details}), and more discussion on the relationship between transformation intensity and unlearning performance from the lens of sample difficulty using the relative Mahalanobis distance (Section \ref{ssec:appendix-discussion-difficulty}).




\section{Proofs} \label{sec:proofs}
\subsection{Proof of Theorem \ref{theorem:two-transform} on Transformation Composition for Unlearning}
\label{ssec:proof-of-theorem}
\begin{proof}
    We prove the first inequality: $I(\theta, \mathcal{T}^{\prime (2)}; D_{forget}) \leq I(\theta, \mathcal{T}_1; D_{forget})$.

    Recall the Definition \ref{def:mutual-info}, the mutual information $I(\theta, \theta^{\prime}; D_S)$ measures the amount of information is shared between models parameterized by $\theta$ and $\theta^{\prime}$ with respect to dataset $D_S$, defined using the KL divergence:

    \begin{equation*}
        I(\theta, \theta'; D_S) = D_{KL}(p_{\theta,\theta'}(x,y) \| p_\theta(x|y)p_{\theta'}(y|x)).
    \end{equation*}
    We have:
    \begin{align}
    I(\theta, \mathcal{T}_1; D_{forget}) &= D_{KL}(p_{\theta,\mathcal{T}_1}(x, y) \parallel p_{\theta}(x|y)p_{\mathcal{T}_1}(y|x)).
    \end{align}
    Since $\mathcal{T}^{\prime (2)} = \mathcal{T}_1 \circ \mathcal{T}_2$ which means $\mathcal{T}^{\prime (2)}(\theta) = \mathcal{T}_2(\mathcal{T}_1(\theta)),$ we also have:
    \begin{align}
    I(\theta, \mathcal{T}^{\prime (2)}; D_{forget}) &= D_{KL}(p_{\theta,T^{\prime 2}}(x, y) \parallel p_{\theta}(x|y)p_{T^{\prime 2}}(y|x)).
    \end{align}

    To establish the inequality, we leverage the data processing inequality for mutual information \cite{cover1999elements}. Consider the Markov chain:
    \begin{align}
    \theta \rightarrow \mathcal{T}_1(\theta) \rightarrow \mathcal{T}_2(\mathcal{T}_1(\theta)) = T^{\prime (2)}(\theta).
    \end{align}
    This Markov structure is justified because $\mathcal{T}_2$ operates solely on $\mathcal{T}_1(\theta)$ and has no direct access to $\theta$. By the definition of function composition, any information flow from $\theta$ to $\mathcal{T}^{\prime (2)}(\theta)$ must pass through $\mathcal{T}_1(\theta)$. The data processing inequality states that for any Markov chain $X \rightarrow Y \rightarrow Z$, we have $I(X;Z) \leq I(X;Y)$ \cite{cover1999elements}. Applying this to our context, we have:
    \begin{align}
    I(\theta, \mathcal{T}^{\prime (2)}(\theta) ; D_{forget}) \leq I(\theta, \mathcal{T}_1(\theta); D_{forget}).
    \end{align}

    Since $\mathcal{T}_2$ is designed to reduce information about $D''_{forget}$ in the model, and the mutual information is evaluated with respect to samples from $D_{forget} = D'_{forget} \cup D''_{forget}$, we have:
    \begin{align}
    I(\theta, \mathcal{T}^{\prime (2)}; D_{forget}) \leq I(\theta, \mathcal{T}_1; D_{forget}).
    \end{align}
    A similar argument applies for the second inequality $I(\theta, \mathcal{T}^{\prime (2)}; D_{forget}) \leq I(\theta, \mathcal{T}_2; D_{forget})$.

    For inequalities 3 and 4 regarding $\mathcal{T}^{''(2)} = \mathcal{T}_2 \circ \mathcal{T}_1$, we establish the Markov chain:
    \begin{align}
    \theta \rightarrow \mathcal{T}_2(\theta) \rightarrow \mathcal{T}_1(\mathcal{T}_2(\theta)) = \mathcal{T}^{''(2)}(\theta).
    \end{align}
    Applying the data processing inequality:
    \begin{align}
    I(\theta, \mathcal{T}^{''(2)}; D_{forget}) &\leq I(\theta, \mathcal{T}_2; D_{forget}), \\
    I(\theta, \mathcal{T}^{''(2)}; D_{forget}) &\leq I(\theta, \mathcal{T}_1; D_{forget}).
    \end{align}
    This completes the proof of all four inequalities in Theorem \ref{theorem:two-transform}.
    \end{proof}

\subsection{Proof of Corollary \ref{corollary} on Monotonic Unlearning Performance}
\label{ssec:proof-of-corollary}
\begin{proof}
The corollary follows by induction.

\textbf{For the base two-transformation case ($m=2$):}
From Theorem~\ref{theorem:two-transform}, we know that when $\mathcal{T}^{(2)} = \mathcal{T}_1 \circ \mathcal{T}_2$ and $\mathcal{T}^{(1)} = \mathcal{T}_1$:
$$I(\theta, \mathcal{T}^{(2)}; D_{forget}) \leq I(\theta, \mathcal{T}^{(1)}; D_{forget}).$$
Similarly, when $\mathcal{T}^{(2)} = \mathcal{T}_2 \circ \mathcal{T}_1$ and $\mathcal{T}^{(1)} = \mathcal{T}_2$, we have:
$$ I(\theta, \mathcal{T}^{(2)}; D_{forget}) \leq I(\theta, \mathcal{T}^{(1)}; D_{forget}).$$

\textbf{For the generalized design of multi-transformation case $m \geq 2$}:
%
Let $\mathcal{T}^{(m+1)} = \mathcal{T}_{_{m+1}} \circ \mathcal{T}^{(m)}$, where $\mathcal{T}_{{m+1}}$ is an additional transformation designed to unlearn some aspect of $D_{\text{forget}}$. Consider the Markov chain: $\theta \rightarrow \mathcal{T}^{(m)}(\theta) \rightarrow \mathcal{T}_{{m+1}}(\mathcal{T}^{(m)}(\theta))$, by the data processing inequality from information theory \cite{cover1999elements}, we know that:
\begin{equation}
    I(\theta, \mathcal{T}_{{m+1}}(\mathcal{T}^{(m)}(\theta)); D_{forget}) \leq I(\theta, \mathcal{T}^{(m)}(\theta); D_{forget}).
\end{equation}
This gives us:
\begin{align}
I(\theta, \mathcal{T}^{(m+1)}; D_{forget}) \leq I(\theta, \mathcal{T}^{(m)}; D_{forget}).
\end{align}

Therefore, by the principle of induction, we have the following inequality for any $m \geq 2$:
\begin{align}
I(\theta, \mathcal{T}^{(m)}; D_{forget}) \leq I(\theta, \mathcal{T}^{(m-1)}; D_{forget}).
\end{align}
This completes the proof.
\end{proof}

\section{More Experiments on CIFAR-10}\label{appd:CIFAR-10}
We provide additional empirical results on the CIFAR-10 dataset to complement the analysis presented in Section \ref{ssec:empirical}. Figure \ref{fig:hypothesis-experiments-cifar10} illustrates these findings.
Figures \ref{subfig:progressive_transform-examples-cifar10} and \ref{subfig:progressive_masking-examples-cifar10} show examples of forgetting samples subjected to increasing transformation intensity via progressive augmentation and masking, respectively.
The corresponding plots in Figures \ref{subfig:progressive_transform_correlation-cifar10} and \ref{subfig:progressive_masking-correlation-cifar10} demonstrate a strong positive correlation between the applied transformation intensity and the overall federated unlearning (FU) performance. These results on CIFAR-10 mirror those observed for CIFAR-100, further substantiating the relationship between increased transformation intensity and improved unlearning effectiveness explored in our work.

\begin{figure}[!ht]
    \centering
    \subfloat[Increasing intensity level of transformation applied to forgetting samples.]{
        \includegraphics[width=0.9\linewidth]{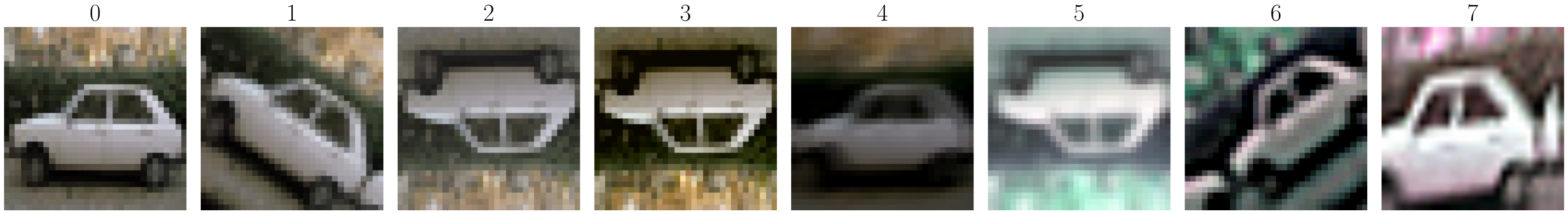}
        \label{subfig:progressive_transform-examples-cifar10}
    }
    \\
    \subfloat[Increasing masking ratio applied to forgetting samples.]{
        \includegraphics[width=0.9\linewidth]{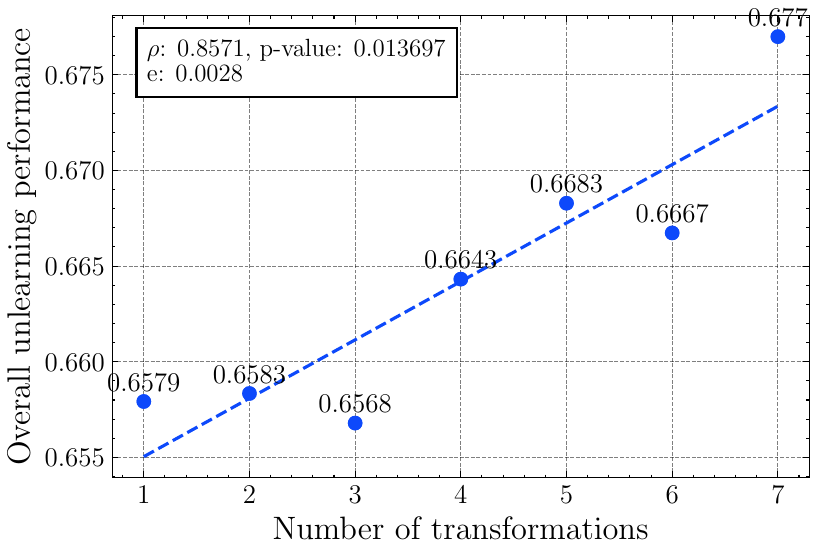}
        \label{subfig:progressive_transform_correlation-cifar10}
    }
    \\
    \subfloat[Overall unlearning performance improves when increasing the intensity level of transformation on the forgetting samples.]{
        \includegraphics[width=0.9\linewidth]{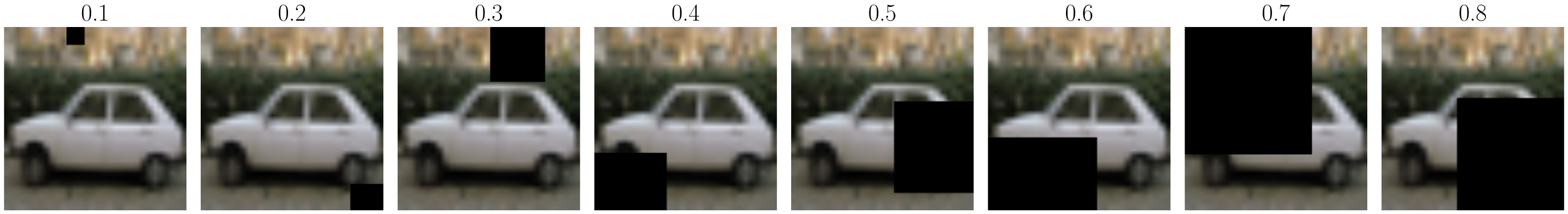}
        \label{subfig:progressive_masking-examples-cifar10}
    }
    \\
    \subfloat[Overall unlearning performance improves when increasing the intensity level of transformation on the forgetting samples.]{
        \includegraphics[width=0.9\linewidth]{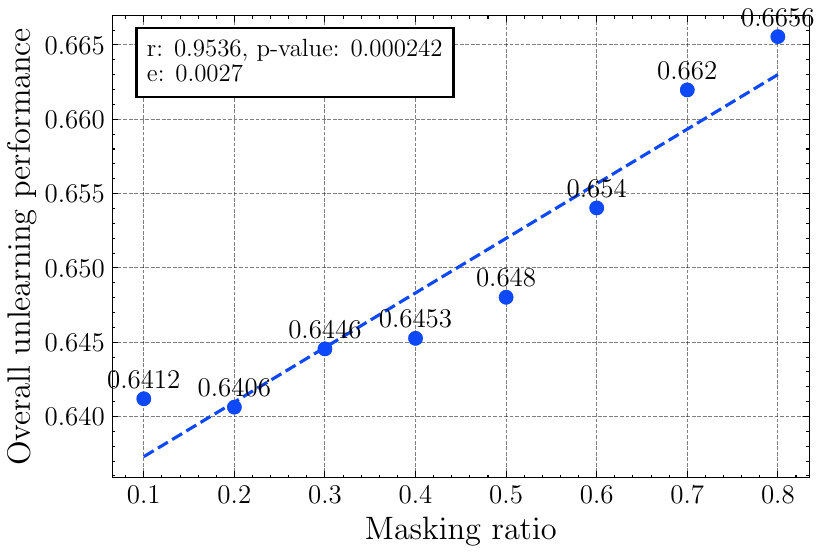}
        \label{subfig:progressive_masking-correlation-cifar10}
    }
    \vspace{0.1cm}
    \caption{Experiments on CIFAR-10 show that the unlearning process becomes more effective when models learn from transformed
sample. (Two Top) Progressive Transformation and (Two Bottom) Progressive Masking. The top figures present forgetting samples with increasing intensity level, while the bottom figures illustrate the correlation between unlearning performance and the intensity level of transformation.}
    \label{fig:hypothesis-experiments-cifar10}
    \vspace{0.5cm}
\end{figure}

\section{Implementation Details}
\label{sec:implementation-details}

\textbf{Metrics.} Three metrics used to evaluate our ToFU framework include:
\begin{itemize}
    \item \textbf{Retain Accuracy}: Measures the unlearned model's performance on the retain sets, calculated as the average accuracy across all client retain sets ($\mathcal{D}_r$).
    \item \textbf{Test Accuracy}: Evaluates the unlearned model's generalization capability on an unseen test set.
    \item \textbf{Membership Inference Attack Efficacy (MIA Efficacy)}: A privacy-focused metric evaluates the success of an unlearning request by determining whether attackers can distinguish forgotten examples from unseen ones. It is defined as the ratio of samples in the forget set that are correctly estimated as non-training samples. Regarding the MIA model, we employ the state-of-the-art method LiRA \cite{carlini2022membership}. The shadow models are constructed by using the global model parameters from the final 5 communication rounds of the FL training process.
\end{itemize}

Local models, ResNet-18 \cite{he2016identity}, are trained for $10$ epochs using Schedule-free-AdamW \cite{defazio2024road} optimizer (batch size $B=128$, learning rate $\eta=3 \times 10^{-4}$).

All experiments were conducted on a system equipped with two NVIDIA RTX 4090 GPUs, an AMD EPYC 9654P CPU, and 192GB of RAM. Our codebase is developed using PyTorch \cite{paszke2019pytorch} and the Flower \cite{beutel2020flower} federated learning framework.
For each client dataset, we reserve 10\% of the client data for validation purposes, ensuring that the validation set reflects the client's local data distribution. Key hyperparameters for the transformations used within the ToFU framework are detailed in Table \ref{tab:transformation-hyperparams}.

\vspace{0.5cm}
\begin{table}[ht]
\centering
\fontsize{7pt}{7pt}\selectfont
\caption{Hyperparameters for Transformations used in ToFU.}
\label{tab:transformation-hyperparams}
\begin{tblr}{
  cells = {c},
  cell{2}{1} = {r=3}{},
  cell{5}{1} = {r=2}{},
  cell{7}{1} = {r=2}{},
  cell{14}{1} = {r=3}{},
  cell{21}{1} = {r=3}{},
  vline{2-3} = {1-23}{},
  hline{1,24} = {-}{0.08em},
  hline{2,5,7,9-14,17-21} = {-}{},
  hline{2} = {2}{-}{},
}
\textbf{Transformation}  & \textbf{Parameter} & \textbf{Value} \\
ShiftScaleRotate         & shift\_limit       & 0.0625         \\
                         & scale\_limit       & 0.1            \\
                         & rotate\_limit      & 0.1            \\
RandomBrightnessContrast & brightness\_limit  & 0.2            \\
                         & contrast\_limit    & 0.2            \\
HueSaturationValue       & hue\_shift\_limit  & 20             \\
                         & sat\_shift\_limit  & 30             \\
RandomGamma              & gamma\_limit       & (80, 120)      \\
RGBShift                 & shift\_limit       & 20             \\
GaussianBlur             & blur\_limit        & (3, 7)         \\
MotionBlur               & blur\_limit        & 7              \\
Downscale                & scale\_min         & 0.25           \\
ColorJitter              & brightness         & 0.2            \\
                         & contrast           & 0.2            \\
                         & saturation         & 0.2            \\
Sharpen                  & alpha              & (0.2, 0.5)     \\
Emboss                   & alpha              & (0.2, 0.5)     \\
GaussNoise               & var\_limit         & (10.0, 50.0)   \\
RandomResizedCrop        & scale              & (0.5, 1.0)     \\
CoarseDropout            & max\_holes         & 1              \\
                         & max\_height        & 0.3            \\
                         & max\_width         & 0.3
\end{tblr}
\end{table}

\begin{algorithm}[t]
\small
\caption{Transformation Function for ToFU}
\label{alg:tofu_transformations}
\SetAlgoLined
\DontPrintSemicolon
\SetKwInOut{Input}{Input}
\SetKwInOut{Output}{Output}
\SetKwComment{Comment}{$\triangleright$\ }{}

\Input{Transformation intensity level $m$}
\Output{Composed transformation function}

\vspace{1mm}
\textbf{Define} TRANSFORMS as an ordered list:\;
\Indp
$\mathcal{T}_1 \gets$ OneOf([HorizontalFlip, VerticalFlip, ShiftScaleRotate])\;
$\mathcal{T}_2 \gets$ RandomBrightnessContrast\;
$\mathcal{T}_3 \gets$ OneOf([HueSaturationValue, RandomGamma, RGBShift])\;
$\mathcal{T}_4 \gets$ OneOf([GaussianBlur, MotionBlur, Downscale])\;
$\mathcal{T}_5 \gets$ OneOf([ToGray, ChannelShuffle, ColorJitter])\;
$\mathcal{T}_6 \gets$ OneOf([Sharpen, Emboss, GaussNoise])\;
$\mathcal{T}_7 \gets$ RandomResizedCrop\;
$\mathcal{T}_8 \gets$ CoarseDropout\;
\Indm

\vspace{1mm}
\textbf{Group transformations by function:}\;
\Indp
Geometric transforms $\gets \{\mathcal{T}_1, \mathcal{T}_7\}$ \Comment*[r]{Spatial modifications}
Color transforms $\gets \{\mathcal{T}_2, \mathcal{T}_3, \mathcal{T}_5\}$ \Comment*[r]{Color space alterations}
Noise/Blur transforms $\gets \{\mathcal{T}_4, \mathcal{T}_6, \mathcal{T}_8\}$ \Comment*[r]{Information degradation}
\Indm

\vspace{1mm}
\textbf{Progressive transformation:}\;
\Indp
$m' \gets \min(m, 8)$ \Comment*[r]{Bound intensity level}
selected\_transforms $\gets$ TRANSFORMS$[0:m']$ \Comment*[r]{Take first $m'$ transforms}
\Indm

\Return{Compose(selected\_transforms)} \Comment*[r]{Apply sequentially}
\end{algorithm}

\section{Details of Applied Transformations} \label{sec:transform-details}
Algorithm \ref{alg:tofu_transformations} presents transformations used in our experiments. These transformations are implemented using the \texttt{albumentation} library.


\section{Additional Ablation Study}
\label{sec:additional-ablation}
\subsection{Ablation study on transformation-invariant regularize strength}
We conduct an ablation study (Table \ref{tab:ablation-regularization}) on the CIFAR-10 dataset to analyze further the impact of the transformation-invariant regularizer strength, controlled by the hyperparameter $\gamma$ (Eq. \ref{eq:loss_function}), on ToFU's performance. We observe that $\gamma=0$ (i.e., removing the regularizer) leads to reasonable performance. However, introducing a small positive regularization ($\gamma=0.05$ or $\gamma=0.5$) slightly improves the overall performance. This suggests that encouraging the model to learn consistent representations across transformations is beneficial. Conversely, setting $\gamma=-1.0$ drastically degrades performance across all metrics. This is expected, as a negative $\gamma$ effectively maximizes the KL divergence in Eq. \ref{eq:loss_function}, forcing the representations of original and transformed samples to diverge, which hinders learning. Increasing $\gamma$ to higher positive values ($\gamma=1.0$ or $\gamma=2.0$) maintains high utility but slightly reduces the MIA Efficacy and the overall score compared to the optimal range. These findings support our choice of $\gamma=0.01$ in the main experiments as a default, although fine-tuning $\gamma$ (e.g., to 0.5 for CIFAR-10) can yield marginal improvements.


\begin{table}[ht]
\centering
\fontsize{7pt}{7pt}\selectfont
\caption{Ablation Study on Regularization Strength.}
\label{tab:ablation-regularization}
\begin{tblr}{
  cells = {c},
  vline{2-5} = {-}{},
  hline{1,8} = {-}{0.08em},
  hline{2} = {-}{},
  hline{2} = {2}{-}{},
}
\textbf{$\gamma$} & \textbf{Test Accuracy} & \textbf{Retain Accuracy} & \textbf{MIA Efficacy} & \textbf{Overall} \\
-1              & 0.1002                 & 0.0945                   & 0.5726                & 0.2558           \\
0               & 0.7730                 & 0.8794                   & 0.2601                & 0.6375           \\
0.05            & 0.7831                 & 0.8811                   & 0.2606                & 0.6416           \\
0.5             & 0.7817                 & 0.8823                   & 0.2643                & 0.6428           \\
1               & 0.7817                 & 0.8838                   & 0.2523                & 0.6393           \\
2               & 0.7873                 & 0.8880                   & 0.2376                & 0.6376
\end{tblr}
\end{table}

\subsection{Ablation study on wall-clock time for training}

\begin{table}[ht]
\centering
\caption{Ablation Study on Wall-clock Time for Training}
\label{tab:wall-clock}
\begin{tblr}{
  width = \linewidth,
  colspec = {Q[508]Q[390]},
  cells = {c},
  vline{2} = {-}{},
  hline{1,11} = {-}{0.08em},
  hline{2} = {-}{},
  hline{2} = {2}{-}{},
}
\textbf{Total Transformations} & \textbf{Running time (s)} \\
0                              & 1277                      \\
1                              & 1458                      \\
2                              & 1543                      \\
3                              & 1591                      \\
4                              & 1655                      \\
5                              & 1745                      \\
6                              & 1902                      \\
7                              & 2003                      \\
8                              & 2027
\end{tblr}
\end{table}

In our framework, the maximum number of transformations is 8, but the actual number of applied transformations will be dynamically determined by the Sample-dependent Transformation Strategy (Section 4). Therefore, the worst case of training time overhead would be 58,7\% and the average case (4 transformations) will be 29.6\%. Although our framework increases training time, it provides (1) 11\% improvement in unlearning performance (Tables 1-2) and (2) 24-28\% reduction in unlearning time (Figure 5), and (3) enhancement of privacy protection against MIA attacks (Figure 4). Moreover, the one-time training overhead is amortized across multiple unlearning operations.

\section{Additional Discussion}
\label{ssec:appendix-discussion-difficulty}

In this section, we discuss the relationship between the transformation intensity level and unlearning performance from the lens of sample difficulty by using the relative Mahalanobis distance (RMD) method \cite{cui2023learning} as a proxy.
RMD scores sample difficulty by computing the difference between class-specific and class-agnostic Mahalanobis distances. While the class-specific Mahalanobis distance measures how typical a sample's features are within its specific class, the class-agnostic Mahalanobis distance measures how common those features are across all classes. A large RMD indicates a difficult sample, meaning it is far from its class-specific mean but close to the class-agnostic mean.
Figure \ref{fig:difficulty_analysis_rmd} shows the relationship between the number of applied transformations and RMD-based sample difficulty on CIFAR-10 and CIFAR-100 datasets via Pearson statistical tests.

This analysis provides a complementary perspective to our framework, offering another insight into why transformation-guided unlearning works. The strong correlation between transformation intensity and sample difficulty (as measured by RMD) suggests that our approach effectively modulates the learning process based on intrinsic data characteristics. As transformations increase, samples become more challenging for the model to memorize, forcing it to learn transformation-invariant features rather than instance-specific details.

\begin{figure}[t]
    \centering
    \subfloat[RMD scores on CIFAR-10.]{
        \includegraphics[width=0.98\linewidth]{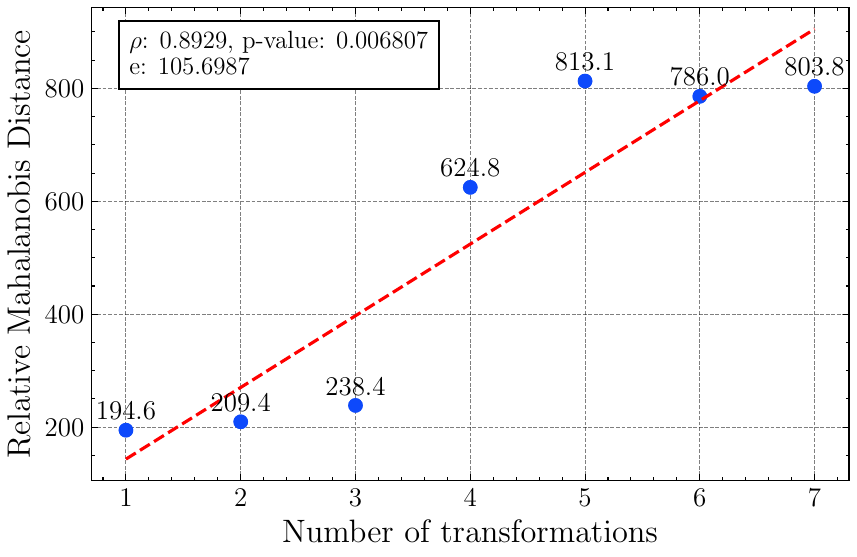}
    }
    \\
    \subfloat[RMD scores on CIFAR-100.]{
        \includegraphics[width=0.98\linewidth]{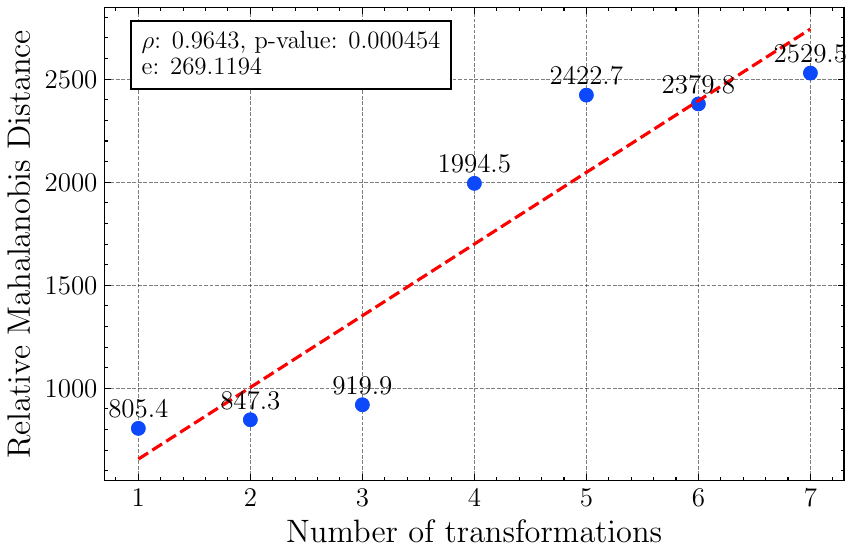}
    }
    \caption{Relative Mahalanobis distance (RMD) for measuring sample difficulty in transformed samples across different levels of transformation intensities.}
    \label{fig:difficulty_analysis_rmd}
    \vspace{0.65cm}
\end{figure}

\end{document}